# Robotic Trail Maker for Rehabilitation in Neurological Conditions: Clinical Use Cases

Srikar Annamraju[†], Harris Nisar[†,*], Dayu Xia[†], Shankar A. Deka, Anne Horowitz, Nadica Miljković, and Dušan M. Stipanović

Srikar Annamraju
srikar@iiitdm.ac.in
Indian Institute of Information Technology
Design and Manufacturing Kancheepuram, Chennai, India

Harris Nisar
nisar2@illinois.edu
University of Illinois Urbana Champaign, IL, 61801, USA

Dayu Xia
dayuxia2@illinois.edu
Zhejiang University, Hangzhou, 310027, China
University of Illinois Urbana Champaign, IL, 61801, USA

Shankar A. Deka
shankar.deka@aalto.fi
Aalto University, 02150 Espoo, Finland

Anne Horowitz
anne.c.horowitz@osfhealthcare.org
OSF HealthCare Outpatient Rehabilitation, Peoria, IL, 61614, USA

Nadica Miljković
nadica.miljkovic@etf.bg.ac.rs
University of Belgrade, Bulevar kralja Aleksandra 73, 11000, Belgrade, Serbia
University of Ljubljana, 25, 1000 Ljubljana, Slovenia

Dušan M. Stipanović
dusan@illinois.edu
University of Illinois Urbana Champaign, IL, 61801, USA

† Indicates co-first authors
* Indicates corresponding author

**Abstract**
Patients with neurological conditions require rehabilitation to restore their motor, visual, and cognitive abilities. To meet the shortage of therapists and reduce their workload, a robotic rehabilitation platform involving the clinical trail making test is proposed. Therapists can create custom trails for each patient, and the patient can trace the trails using a robotic device. The platform can track the performance of the patient and use these data to provide dynamic assistance through the robot to the patient interface. Therefore, the proposed platform not only functions as an evaluation platform but can potentially train the patient in recovery. Further, neural networks are trained to classify between patients and healthy subjects and to forecast their movements using the data collected. The developed platform has been validated at a rehabilitation center, with therapists and patients operating the device. It was found that robotic assistance provided improved performance accuracy and increased the speed at a significant level amongst patients. The usability of the developed platform was demonstrated in a clinical setting to simultaneously assess the patients' condition and intelligently assist them in recovery.

## 1 Introduction

Neurological conditions such as stroke, Parkinson's disease, and multiple sclerosis, are a major concern across the globe, with negative effects on quality of life and overall lifespan [1], [2]. Typically, neurological conditions disrupt brain function by blocking communication and/or damaging brain cells. This translates to a decline in patient's motor, vision, and cognitive abilities [3].

To regain these abilities and help improve the survivor's quality of life, rehabilitation is often delivered in specialized clinics [4], [5], [6].

Evaluation methods like Fugl-Meyer (FM) [7], O'Connor Finger Dexterity Test, 9 Hole Peg [8], and Minnesota Rate of Manipulation Test [9], as well as treatments like mirror therapy have been in place since decades to address the neurological conditions. While the rehabilitation techniques themselves are unequivocally proven to be a potent solution, imparting them is becoming challenging, given the increasing workload of therapists [10], [11]. With the number of survivors from neurological conditions rising each year [12], both occupational and physical therapists have to handle additional work loads. In addition, a constant concern for the therapists is sustaining the motivation levels of the patients [13], [14], not only to carry on with their daily lives, but even to participate in the therapy sessions.

Robotic rehabilitation has been proposed to provide patients with consistent and trackable therapy while easing the therapists' task [15], [16]. Particularly, over the past two decades, several commercial upper-extremity rehabilitation robots, such as Myomo e100 (Myomo Inc., 2007), CyberGrasp (CyberGlove Systems, 1998), InMotion HAND (Interactive Motion Tech., 2007), PowerGrip (Broaden Horizons Inc.), ReoGo (Motorica Medical Inc.), and Reha-Digit (Reha-Stim, 2008), have been developed. The clinical effects of incorporating robots for post-stroke rehabilitation are discussed in [17].

The rehabilitation robots proposed varied in their design and have different form factors, such as exo-skeleton robots (Salford, ESTEC, IntelliArm, etc.) and end-effector based robots (MIT-Manus, ACRE, HandCARE, etc.), each with their distinct features. For aiding the patient, various types of equipment, such as active devices (those which assist), passive devices (those which resist), and haptic devices (those which render customized forces) are developed. A detailed survey of rehabilitation robots, particularly pertaining to upper-extremity, can be found in [18]. However, the existing robotic platforms have the following limitations:

i. Considerable number of existing platforms provide generalized assistance in limb movements but are expensive to develop and implement [19].
ii. While robots may assist the patients, there is no platform to the best of our knowledge, which can simultaneously assess the patient's performance and assist their recovery.
iii. The existing robots have very little ability to intelligently predict and learn the patient's needs.

The objective of this work is to present the development of a robotic rehabilitation platform, which can evaluate and accordingly assist the patient, targeting specific therapy outcomes.

Motivated by our previous results of a robotic mirror therapy platform using an off-the-shelf robot [20] and positive feedback from patients and medical staff, an advanced version of the robotic platform for carrying out the 'trail maker' therapy is presented here.

Trail maker is a conventional task for patients with neurological conditions [21], where they are asked to connect dots in a specific order on a sheet of paper (thus making "trails"). Through this task, the therapist observes the patient's vision and cognitive abilities to first identify the numbered sequence of dots, and then their upper-extremity motor abilities for smoothly connecting them. Thus, trail maker is typically an evaluation test performed to assess the patient's specific abilities [22].

In the robotic trail maker (RTM) developed in this work, several additional features are incorporated compared to the traditional trail maker, enabling it to function not just for evaluation purposes, but extending it to a therapeutic training technique. The addition of features such as flexibility in choosing the trails, dynamically varying assistance, enabling task repetition over multiple sessions, and performance analysis, leads to (i) the therapist being only minimally involved in the session, and (ii) an interactive experience for the patient which sustains their motivation.

In addition to these features, Deep Learning (DL) techniques are explored to classify patients and predict their trajectories during therapy, so that appropriate assistance may be provided. In recent years, DL is being extensively exploited in the context of enhancing rehabilitation delivery. One of the first approaches to use DL in assessing rehabilitation performance can be seen in [23], where joint displacement data of body parts are processed for various exercises. In [24], an attention-guided hierarchy is presented which can continuously score the patient for exercises performed without the clinician. A DL technique specifically targeting upper extremity motor recovery after stroke is presented in [25], which uses Long Short-Term Memory (LSTM) neural networks to evaluate the training movements. Other DL approaches for scoring the patients economically (without usage of additional sensors) are found in [26], [27]. However, since a majority of these approaches intend only evaluation, an attempt is made herein to also exploit DL for predicting the patient's trajectories, a critical preliminary step in providing real-time robot assistance to the patients. Thus, the therapists report the proposed RTM as a valuable and beneficial platform serving the purposes of therapy and evaluation.

The developed platform is unique in exploiting the features of a haptic device for specific rehabilitation methods. Upon development and incorporation of safety features, and with appropriate IRB approvals, the RTM has been tested first with 10 healthy subjects, and then with 11 patients affected by neurological conditions at the Outpatient Rehabilitation Center of OSF HealthCare, Peoria, IL, USA. Multiple occupational therapists were trained on the RTM platform, who use the developed RTM as a regular means of therapy for their patients. The protocol used for data collected from both the healthy subjects and patients, along with the inferences from the data, are detailed in later sections of this paper. In summary, the following are the technical contributions of this work:

i. Developed a trail maker platform using a low-cost commercial robot, that is convenient and safe to use by therapists and patients.
ii. Collected robotic data from patients and healthy subjects and analyzed the data to validate the platform's ability to assist and evaluate patients in a preliminary study.
iii. Trained deep learning models to classify healthy subjects from patients and for trajectory forecasting and validated their performance.

# 2 Development of the RTM Platform

The RTM platform is designed and developed such that it can be switched amongst different patients with practically no additional setup time. The only startup time required would be if the new subjects operating the device need any practice to get familiar with the platform, typically for a period of few minutes.

## 2.1 Choice of Device

The Touch haptic device from 3D Systems®, which is a 6 Degrees of Freedom (DoF) end-effector type device, is chosen for developing RTM. The device footprint being as small as ~(17 × 20) cm with weights of only ~1.4 kg, makes it a convenient table-top option for installation and portability in rehabilitation clinics. Moreover, the low-cost nature of this device (<$5k) increases its potential for adoption in therapy clinics. The device constraints on limited workspace and force rendering capabilities are known to be concerning for multiple applications [28]. However, for the purpose of recreating trail making, its planar workspace of ~(43 × 35) cm is larger than the typical US Letter size paper used in traditional trail making, making it sufficient for this application. Also, during trail making, the user does not have to interact with virtual objects (other than the paper) and thus its force feedback capability of 3.3 N is sufficient to provide assistive forces to the user's arm, which are further detailed in Section 2.4 Further, the robot is not meant to support the entire arm of the patient and interacting with the platform can be done with the arm rested on a surface.

## 2.2 User Interface

For interfacing the haptic device in real-time, Unity® software, which facilitates a highly user-friendly interaction environment, has been chosen. With the plugins developed for the device in Unity, it is possible to read the instantaneous positions and end-effector forces, as well as control the robot to reach specific positions.

An RTM environment is created where the therapist first chooses to register a new patient, or pulls the prior sessions records of an existing patient with their assigned ID. For each session, a whiteboard appears on the screen in front of the patient, resembling a traditional trail maker scenario. As the patient holds the robot end-effector and makes upper-extremity movements, a virtual pen on the screen emulates corresponding motions.

## 2.3 Therapist-centric Features

### 2.3.1 Dynamic Trail Making

Upon the whiteboard of the RTM interface, the therapist can add any number of 'Targets (T)', which are programmed to appear in a numbered fashion, and move them on the screen. Fig. 1 shows a set of 4 connected targets in red, and the virtual pen that corresponds to the position of the stylus of the robot (end-effector).

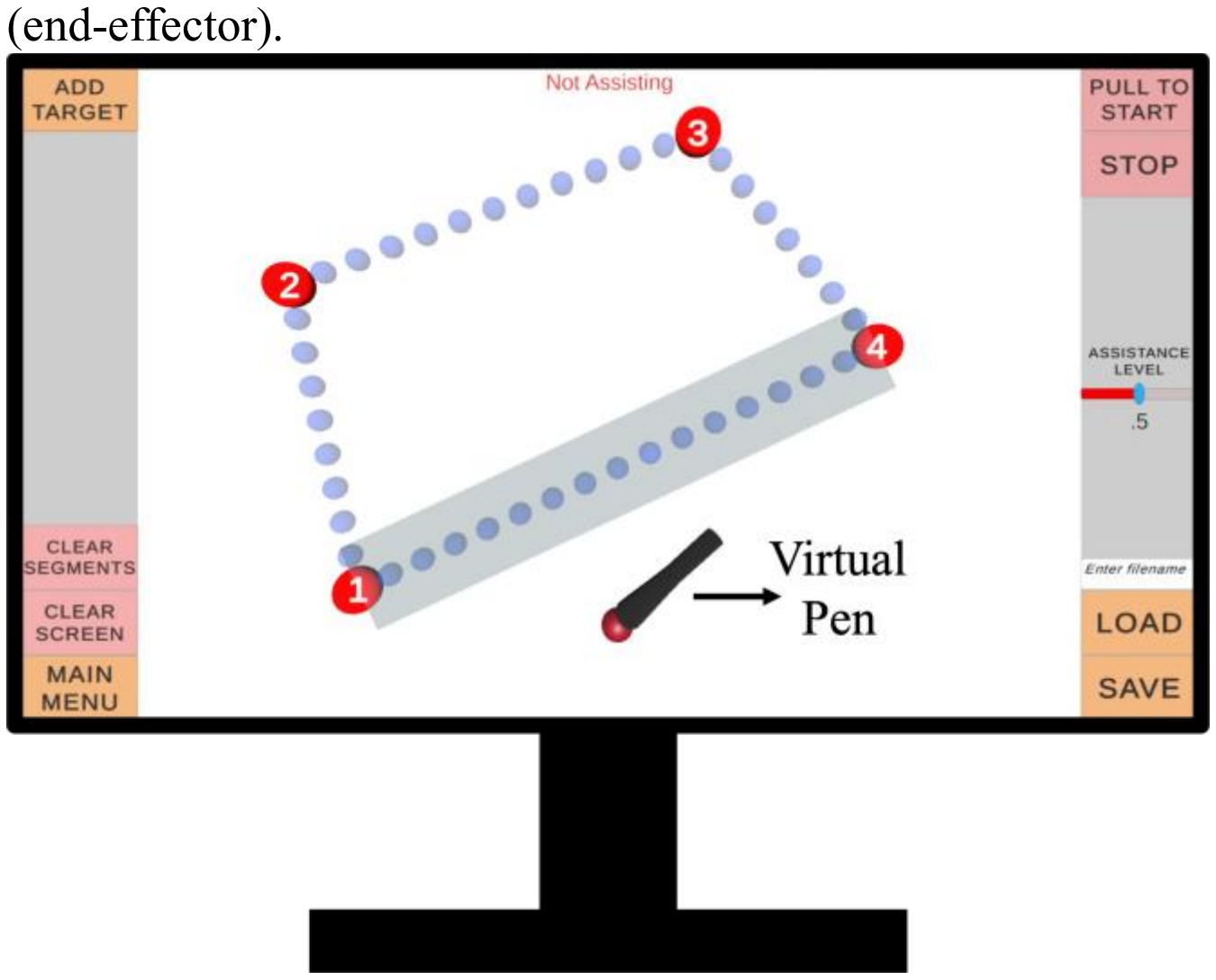

**Fig. 1** RTM whiteboard with sample set of 4 targets

Additional targets can be added through the 'Add Target' button. Arranging the targets in the desired shape forms the required trail. The 'Looping' feature ensures that the last target is connected back to the first, so that the patient can repeatedly follow the same trail. This feature becomes crucial for both therapist and patient from the viewpoint of assistive forces, which will be discussed in the following sub-section. Even after the segments are generated (shown by the blue dotted lines Fig. 1), the therapist can move the targets, and the segments would be reconnected automatically.

From the clinical perspective, the choice of trail for a particular patient plays a key role. Patients with different neurological and musculoskeletal conditions develop specific movement disability and need to practice making trails accordingly [29]. For instance, stroke patients with spasticity suffer from muscle stiffness and need smoother trails (such as a circle) in a continuous loop to restore free muscle movement. On contrary, patients with flaccid paralytic condition suffer from sudden limb weakness and lower eye coordination [29]. In such cases, trails with sharp corners (such as alphabets 'A', 'M', or triangle) are required for their recovery. When appropriate trails have been identified, they can be saved and loaded in future sessions. This feature allows for personalized delivery of rehabilitation which is important for patient motivation [30]. With the patient-specific trails created/loaded, the therapist begins the session. The buttons for 'Save', 'Load', 'Begin' etc. can be seen in the interface shown in Fig. 1.

### 2.3.2 Patient Score Tracking

At the beginning of each session, clinical metrics such as FM scores of the patient will be displayed from the previous session. If the therapist finds that the patient's recovery metrics have changed, these scores can be updated. The FM score integration into the RTM helps the therapist to track the patient's progress over a series of sessions. Such history of scores also helps the platform to automatically choose the reference trail for the patient's future sessions (detailed in Section 5).

### 2.3.3 Adaptive Assistance

While the patient attempts to follow the trail with their robot, the robot provides an assistance force. The assistive force is a spring force pulling back the user towards the trail in proportion to their degree of deviation. The magnitude / spring constant of this force can be optionally scaled by the therapist based on the patient's abilities through a slider. The therapist sets the slider-based assistance dynamically even during the same session, since within a given trail, sharp corners may demand higher mobility which the patient may not be able to naturally exert. Recent works [31] provide guidelines on how to prioritize the parameters to make the platform intelligently assist based on patient's need. The proposed method also adapts the assistance based on the instantaneous requirements.

### 2.3.4 Data Recording

After the patient goes through a certain number of loops for a trail, the therapist may clear the screen to create new trails within the same session. After a sufficient number of trails have been completed, the therapist ends the session and the complete session data, including a) the target positions on each trail, b) the timestep (sampling frequency of 30 Hz), c) the patient's position at each instant, d) the assistance level needed at each instant, and e) the video recording of the screen during the entire session, are stored.

## 2.4 Patient-centric Features

### 2.4.1 Startup Calibration

To start the process, if new users of RTM find it difficult to calibrate measurements or identify themselves on the screen, the option of 'Pull to Start' (seen in Fig. 1) automatically brings the physical robot to the first target position, and the virtual pen aligns accordingly. Since further assistance is almost always required, even after device calibration, three layers of assistive forces are supplied to the patient, to facilitate the patient in accurate trial following, and consequently in exercising required joint motions.

### 2.4.2 Plane Retaining Force

Since the physical robot is capable of moving in 3D space, but the trail maker functions only on a selected 2D plane (shown in green in Fig. 2), the patient's motion needs to be constrained in the plane. While 3D movements and practicing Activities of Daily Living (ADL) are important in rehabilitation and are thoroughly detailed in the previous work [29], this work aims at specific 2D movements of trail making, since it is focused on achieving specific therapy outcomes. For constraining movements to the plane, a force which always drags the user into the plane (shown in brown in Fig. 2) is supplied throughout the session, irrespective of the therapist's commanded assistance level. When the patient accidentally moves out of the plane, the color of the virtual pen changes and the plane retaining force gets activated.

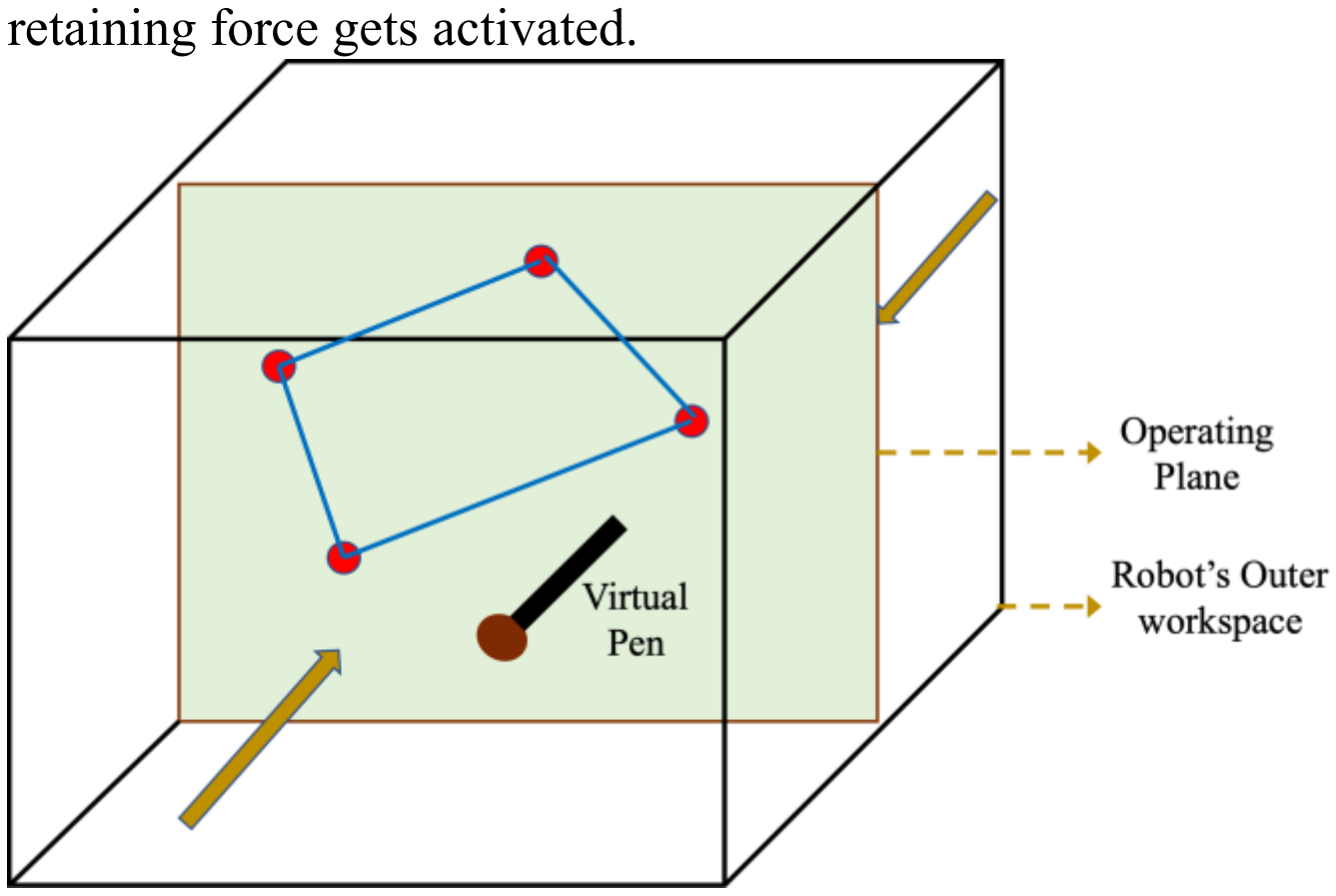

**Fig. 2** Forces constraining patient to the trail's plane

### 2.4.3 Path Leading Force

While the patient stays within the trail plane, their actual task of following the reference can still be challenging. Patients often deviate from the defined path, and this is when the therapist activates the assistance through a keyboard input. When activated, the path leading force assists the patient in a two-fold manner: a) by pulling the patient towards the nearest point (*Np*) on the reference trail and b) by pulling the patient towards the next target (both visualized as brown vectors in Fig. 3). A resultant of these two vectors acts upon the patient's stylus, thus bringing them closer to the trail, and simultaneously ensuring their forward motion along the trail. The magnitude of assistance is made directly proportional to the degree of deviation, with the proportionality constant set by the therapist using the slider (shown in Fig. 1).

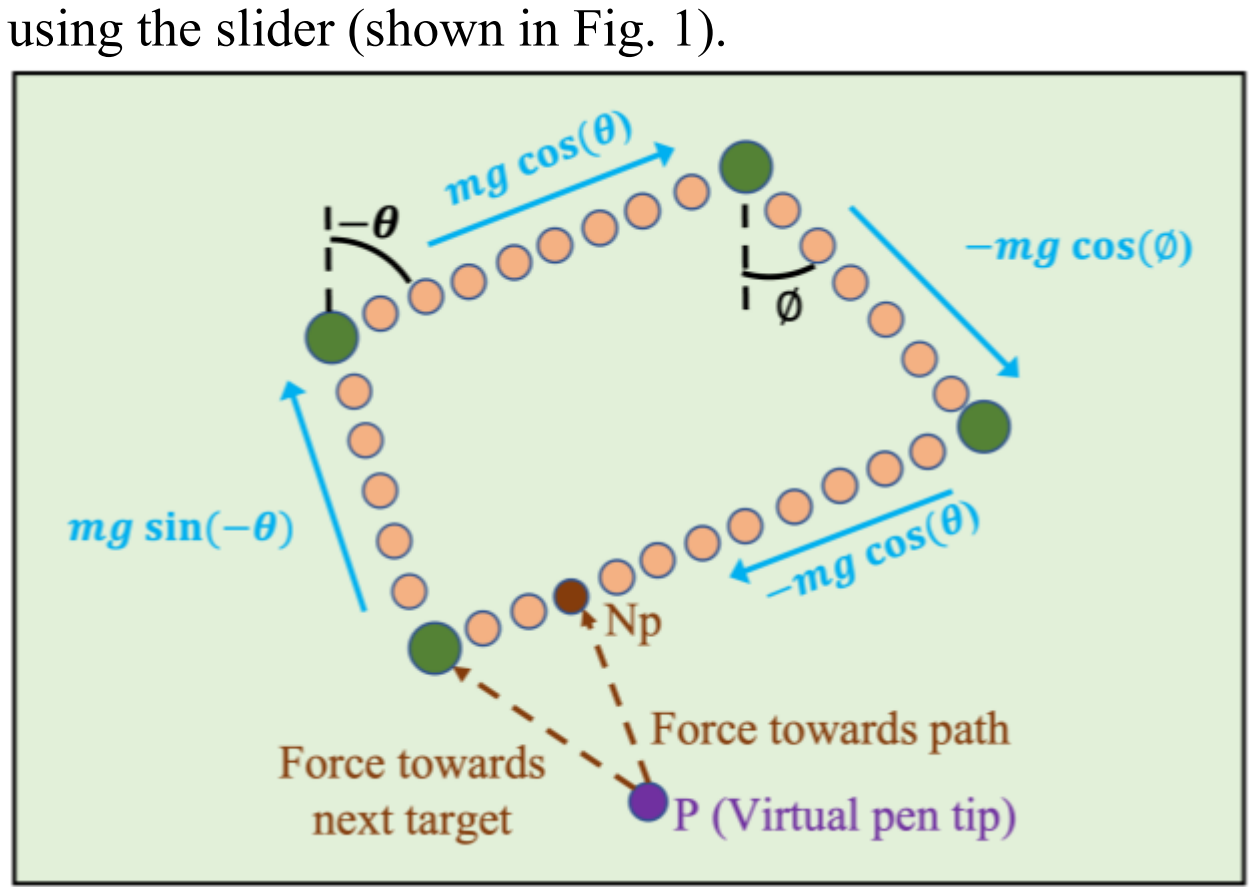

**Fig. 3** Path leading and gravity compensating forces

For smoother assistance, several points are generated between each consecutive targets by linearly interpolating pairs of subsequent target points, $T_i(x_i, y_i)$ and $T_{i+1}(x_{i+1}, y_{i+1})$. The number of points on each segment can be adjusted through a spacing variable $(s)$, set to 8 mm by default, and acts as a resolution parameter. A parameter $n_i$ is defined as the number of intermediate points generated for each $T_i$ and $T_{i+1}$ as

$$n_i = \text{floor}\left[\frac{\text{dist}(T_i, T_{i+1})}{s}\right] \quad (1)$$

The coordinates of the interpolated points $(x_k, y_k)$ between the two targets are computed as,

$$x_k = x_i + k * s * \cos\big(\text{atan2}(y_{i+1} - y_i, x_{i+1} - x_i)\big), \quad (2)$$

$$y_k = y_i + k * s * \sin\big(\text{atan2}(y_{i+1} - y_i, x_{i+1} - x_i)\big), \quad (3)$$

$$\forall\, k \in \{1,2,3 \ldots .\, n_i\}.$$

At each instant, the nearest point is searched by computing the distance of the virtual pen (located at $(x, y)$) from each point on the nearest segment and identifying the shortest distance $(d_s)$ computed as

$$d_s = \min_k(\|(x - x_k, y - y_k\|_2) \quad (4)$$

### 2.4.4 Gravity Compensating Force

Since the patient visualizes the trails on a vertical screen, the most intuitive plane of operation is vertical. This leads to a situation where the patient is subjected to gravitational forces during trail making, which is likely to be absent in the traditional trail maker, due to the paper being in the horizontal plane in most cases. Counteracting gravity provides the patient with a very natural experience of freely moving the stylus, with the same amount of force causing the same motion profile irrespective of the direction of motion.

To enable gravity compensation, the angle which each segment vector creates with the vertical axis is computed (from the known target positions), and the component of the gravity vector along the segment is calculated. For the sample trail, the magnitudes of the gravity compensation force along each segment are depicted in Fig. 3. Another major advantage delivered to the patient through the gravity compensating force is that the mass of the robot stylus is obscured. This feature is particularly important for any robotic platform for those with neurological conditions, since overcoming inertia and getting started with motion is a common concern.

The steps that occur in the proposed RTM are summarized in a sequential manner in Fig. 4.

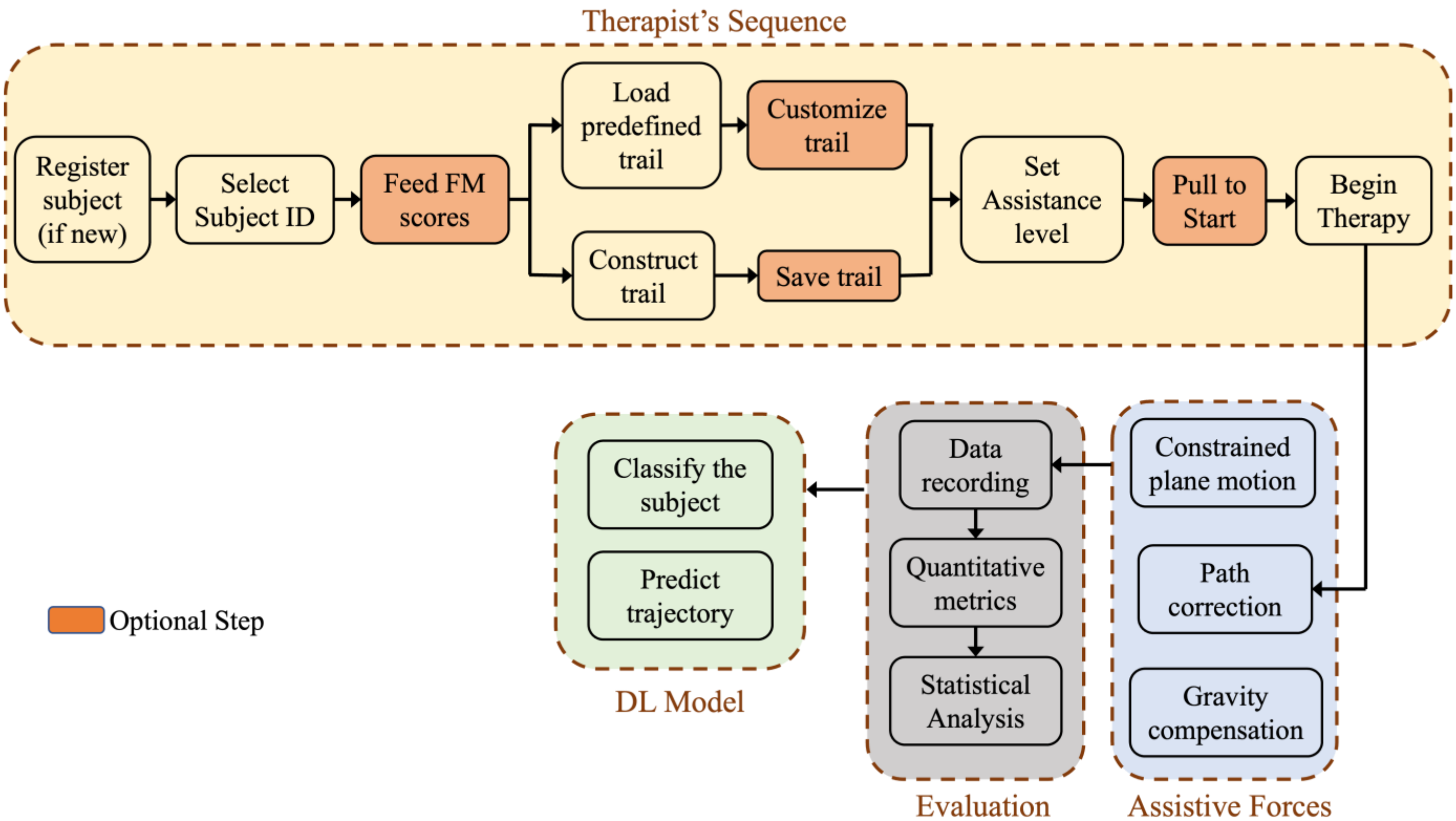

**Fig. 4** Sequential steps of the proposed RTM (these steps can be repeated for the same patient within a session)

### 2.5 Technical Details

It is observed, from preliminary investigation, that patients with cognitive disabilities, especially those with little experience in operating gadgets, might have difficulty in locating themselves on the computer screen, and follow the movements of their hand [20]. To enable easier localization, not only the tip of the virtual pen is highlighted, but also the segment on the trail which the patient is the closest to is highlighted with a translucent boundary (Fig. 1). This significantly reduces the search space for any user, *i.e.*, instead of searching through the whole monitor screen, they tend to look around the highlighted zone, which is typically <10% of the screen area.

To ensure complete safety and comfort for the patients, high enough resolution between targets (discussed in 2.4) is identified through the experiments and standardized. While pulling the user towards a target, the direction vector is firstly adaptively computed, and the magnitude is normalized to eliminate impulsive forces. The normalized forces for each direction are given separate control over the magnitude by decoupling them in planar and perpendicular directions.

Finally, to make the user experience more intuitive, all the buttons on the interface are activated when they are applicable. For instance, the 'Generate Segments' button will only be displayed when there are at least two targets added, and the trail has at least one unconnected segment.

### 2.6 Safety Measures

The developed RTM is tested by the engineers considering various edge cases, such as the user suddenly letting go of the device, the patient being stuck at a corner point, and the platform providing inappropriate assistance level. To ensure that the users get a quick and comfortable learning curve with the platform, the interface is developed considering the novice users of technology.

The device itself offers a sound safety layer, because of it being an end-effector based (not physically tied to the patient's arm), causing no harm to the operator even if abruptly left alone. However, to ensure further layers of safety, software mechanisms of velocity and torque saturation are added to the RTM. For typical trail making, the average velocity of the robot joint motors during RTM is about 15 cm/sec, after accounting for the gear ratios. So, a threshold of 40 cm/sec, which even stroke patients will be able to conveniently handle is set for the magnitude of joint velocity. If by any malfunction the robot tends to fly away from the operator's hand, then the instantaneous velocity of the joints would spike, and all the motors are programmed to be disabled in such cases. Similarly, if the robot for any reason exerts joint torques greater than the patient's admissible range (~2 N observed for RTM) while providing the assistive forces, all the robot motors are programmed to be disabled (shutdown).

## 3 Experimental Protocol

### 3.1 Experimental Setting

With the demonstration of the RTM utility and the safety protocol, IRB approval has been obtained (IRB #00000688)

from the Peoria Institutional Review Board, to carry out human subjects' research at the OSF HealthCare St. Francis Medical Center, Peoria, IL, USA. Firstly, therapists at the rehabilitation clinic of this medical center, whose active inputs were incorporated in the RTM, were trained to use the RTM. All subsequent human subjects operating the RTM happened only under the supervision of the trained therapists and all subjects provided their informed consent to participate in the study. A subject operating the RTM platform can be seen in Fig. 5.

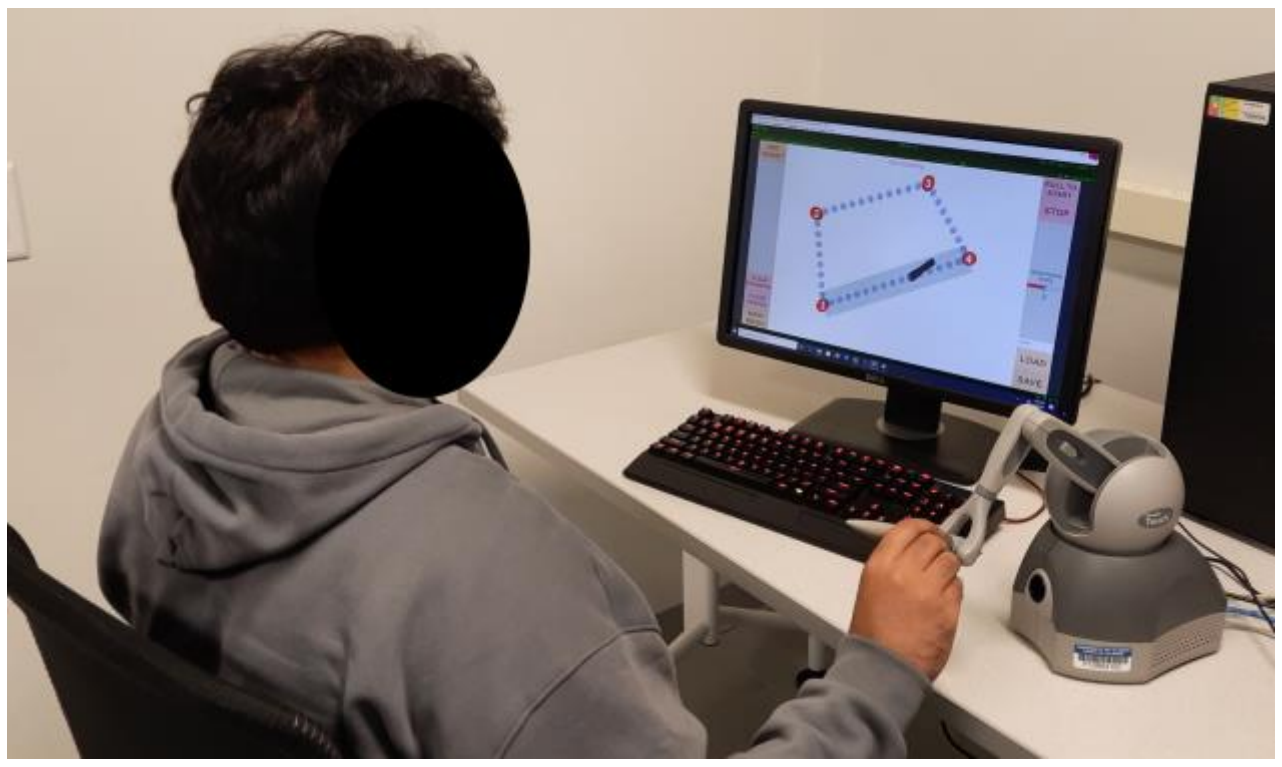

**Fig. 5** Human subject operating RTM platform

### 3.2 Study Sample

While RTM is intended for patients' use, it is necessary to enable the healthy subjects to operate and test the platform with the aim of providing feedback, both quantitative and qualitative. The data from healthy subjects can be used as a benchmark to analyze how closely patients' performance (with assistance) compares with healthy subjects' manipulation. Moreover, with deep learning techniques being developed to enhance patient's experience, the data from healthy subjects could be used as a reference. Thus, a total of 10 healthy subjects and 11 patients are included for the study. The inclusion criteria are as follows:

(i) All the subjects (whether healthy or patients) are chosen in an appropriate age group of (18-60 years). This is to eliminate the potential inaccuracies caused by children or the elderly, while getting accustomed to new technology.

(ii) For the patients, only those whose naturally dominant hand is affected by a neurological condition are considered for the study, for a fair initial comparison. The platform is still valid for other patients, but for data collection purposes and unbiased analysis (that is not affected by dominant hand performance), these inclusion criteria has been chosen.

The demographics of the patient population and healthy subjects are reported in Table 1 and Table 2 respectively. For the patients, an evaluation is performed by the therapist to score their strength in each joint of the dominant arm, as it conveys a significance in providing assistance accordingly.

**Table 1** Demographics and characteristics of patients

| | |
|---|---|
| Age in Years (mean ± standard deviation) | 40.9 ± 6.2 |
| Number of Male Participants | 7 |
| Number of Female Participants | 4 |
| Number of Stroke Survivors | 6 |
| Number of Spinal Cord Injuries | 2 |
| Number of Traumatic Brain Injuries | 2 |
| Number of Parkinson's Disease Diagnoses | 1 |
| Time affected by neurological condition (mean ± standard deviation) | 10.1 ± 2.7 months |
| Average shoulder flexion strength | 3.33 / 5 |
| Average Shoulder Extension Strength | 3.77 / 5 |
| Average Elbow Flexion Strength | 4.25 / 5 |
| Average Elbow Extension Strength | 3.75 / 5 |
| Average Wrist Flexion Strength | 3.83 / 5 |
| Average Wrist Extension Strength | 4.0 / 5 |

**Table 2** Demographics of healthy subjects

| | |
|---|---|
| Age (mean ± standard deviation) | 29.2 ± 4.1 years |
| Number of Male Participants | 4 |
| Number of Female Participants | 6 |

### 3.3 Data Collection

A set of 5 trails – Hexagon shape, Triangle shape, Circle shape, '∞' symbol, and Alphabet 'B' are chosen as references in agreement with the therapist and preliminary recordings. The rationale behind the choice of these references comes from combining smooth curves and sharp edges, given the pathological conditions of the subject population.

The patient goes through each trail in the following order:

**1)** ***No assistance mode:*** 10 loops with assistance always OFF (note that plane retaining and gravity compensation forces are always active, and only the path leading forces are set to zero). In effect, without the assistance in following the trail, this mode is like the traditional, non-robotic trail making; and

**2)** ***Continuous assistance mode:*** 10 loops with assistance always ON (path leading forces set to a scale of 1). The protocol along with its consequences is depicted in Fig. 6.

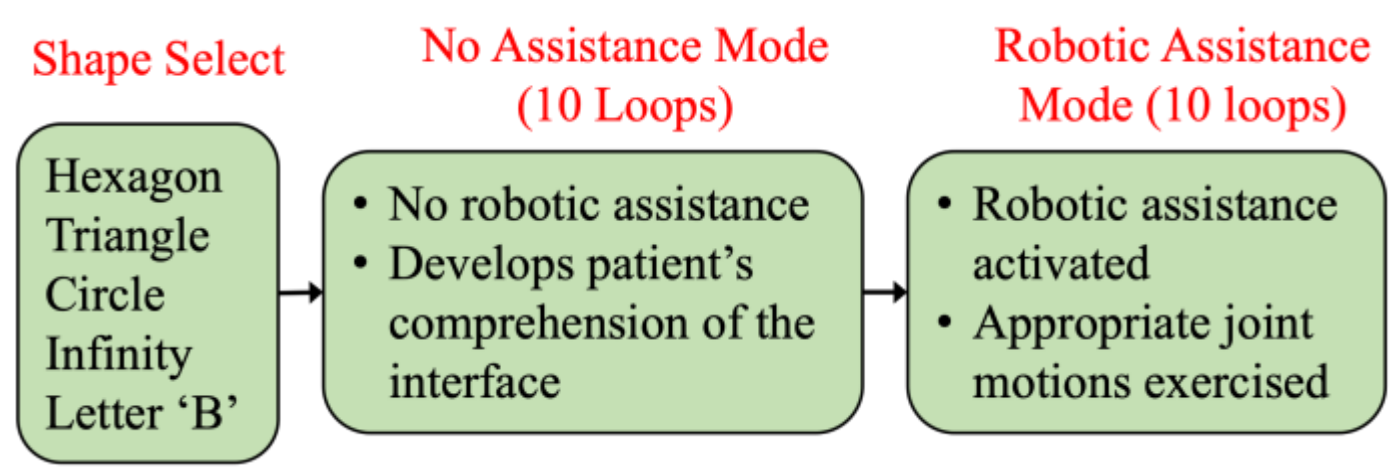

**Fig. 6** Data collection protocol – various modes of RTM

Healthy subjects only completed each trail without assistance. It was observed that assistance made the task more difficult for healthy subjects because they did not need help in completing the task and the assistance acted more as a disturbance for this group.

## 4 Human Subject Data and Inferences

Although RTM has been validated with 11 patients and 10 healthy subjects, in this section, the case study of one representative patient and one healthy subject are detailed. Also, statistical data from all the subjects are presented. The patient in this case study is affected by a spinal cord injury for about 10 months, consequently getting the right arm partially paralyzed. This subject has difficulty in gripping and moving moderate daily life weights, such as a water bottle and has been going through traditional physical and occupational therapy for about 30 sessions (20 minutes per session), some of which included the traditional trail maker. However, since traditional trail maker was only carried out for patient evaluation, the patient found the objective of developed RTM more involving.

The patient is introduced to RTM with arbitrary demo trails and then taken through the protocol described in Section 3.3. The trajectories made by the patient while following the 5 reference trails without and with assistance are shown in Fig. 7 (a & b respectively). The patient's accuracy is visibly enhanced with robotic assistance, especially in complex shapes, like 'B'. It can be observed that though the patient is significantly precise in tracing 'B' during all 10 loops, their accuracy is quite low, visible from a uniform deviation. This characteristic is commonly seen in cases of spinal cord injury, and the robotic assistance is able to improve accuracy.

To validate the degree to which the patient's assisted performance has come closer to the healthy condition, similar protocol is followed with a male healthy subject (right hand dominant), and their trajectories are reported in Fig. 7(c).

To quantitatively study the degree of deviation caused by the subject, the nearest point of the pen on the reference trail is computed at each time instant ($d_s$). This metric yields the instantaneous deviation of the subject, and an average over the session provides the average deviation ($D_{avg}$) of the subject computed as

$$D_{avg} = \frac{\sum_{t=1}^{end} d_s}{n_t} \tag{5}$$

where $n_t$ is the number of timesteps.

The average deviation for the patient across all the shapes improved from 4.3 mm to 3.4 mm (median of 3.1 mm), which is almost the same as healthy subject's average deviation of 3.5 cm (median of 3.2 mm). In addition to $D_{avg}$, a speed metric ($s$) is also computed to study the performance which is not visually captured through the trajectory tracing shown in Fig. 7 and is computed as

$$s = \frac{\mathrm{p_t} * n_l}{t_s} \tag{6}$$

where $\mathrm{p_t}$ is the perimeter of the trail, $n_l$ is the number of loops made and $t_s$ is the session duration.

Similarly, the patient's average speed across all shapes improved from 26.5 mm/sec to 47.7 mm/sec (median of 54.2 mm/sec), again being almost close to the healthy subject's average speed of 46.7 mm/sec (median of 43.1 mm/sec).

To obtain a comprehensive analysis of the study, the box plots of these two metrics for all the 21 subjects are presented in Fig. 8.

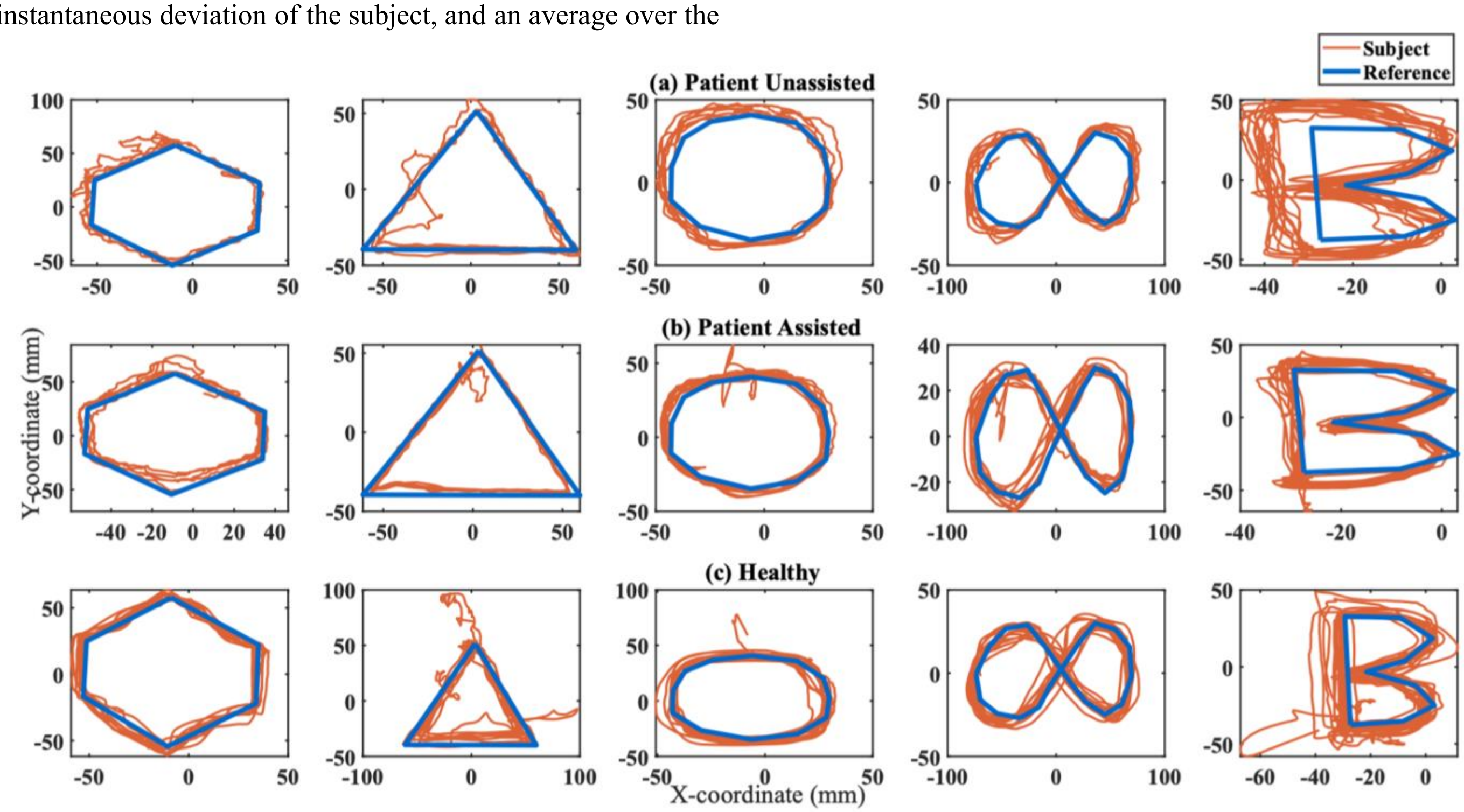

**Fig. 7** RTM data from the single patient and healthy subject case study

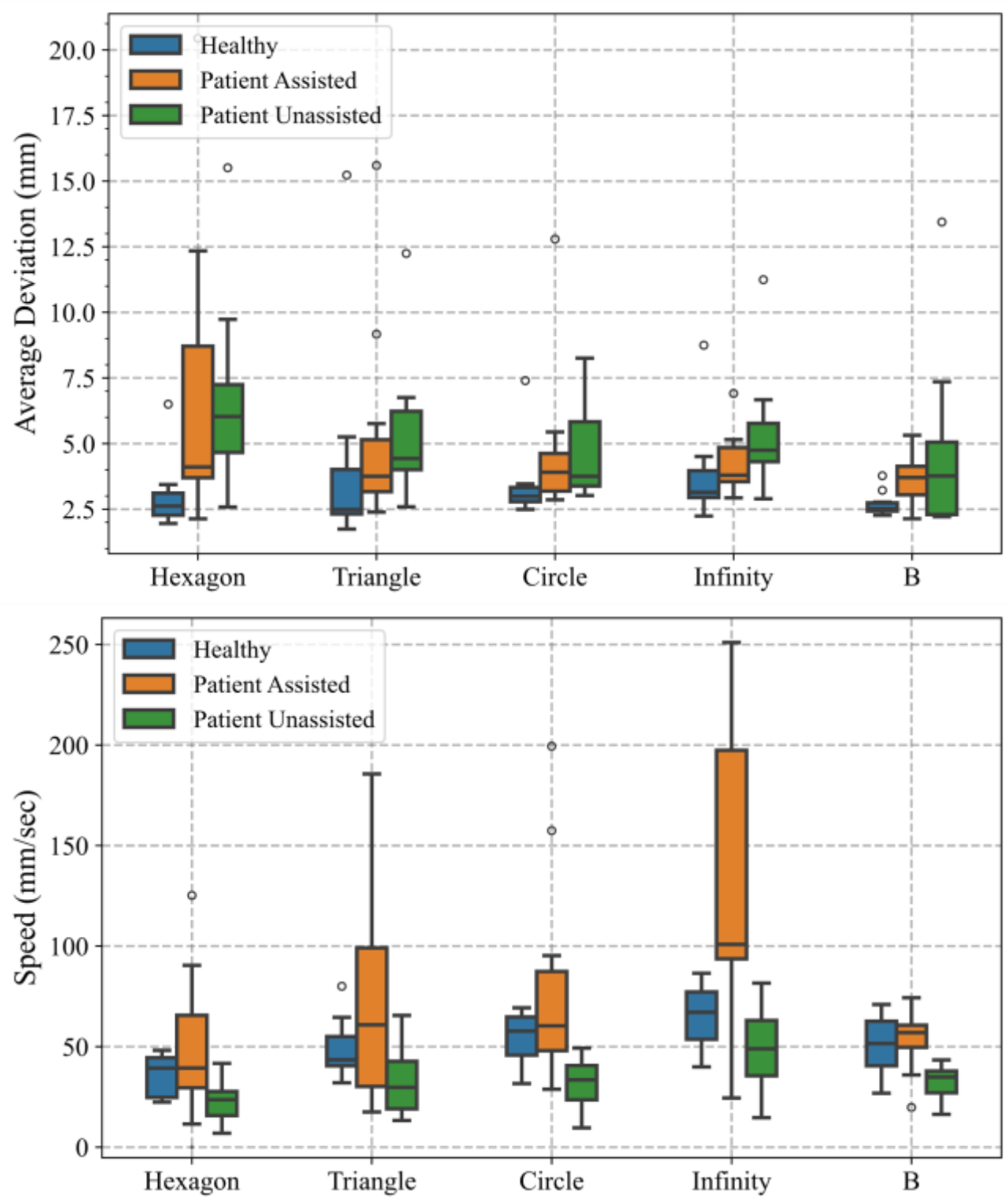

**Fig. 8** Box plots of average deviation (top) and speed (bottom) by each shape for all participants

The distribution of average deviation and speed for the three groups of subjects (Healthy, Patient Assisted, and Patient Unassisted) is tested with the Shapiro–Wilk tests [32]. Since the data distributions were not normal, the non-parametric Kruskal-Wallis tests [33] for multiple comparisons of average deviation and speed are used. Dunn's tests were employed for post hoc comparisons between groups of subjects [34]. For all tests, significant level of $p$-value was set to 0.05.

Analysis revealed a significant difference in average deviation between healthy subjects and patients without assistance ($p < 0.001$). This result provides evidence that the platform behaves like the pen and paper trail maker exam where impairment level is related to performance. However, there was significant difference between healthy subjects and patients with assistance ($p < 0.001$) and there was no difference between patients with and without assistance ($p = 0.458$) when it came to average deviation. Fig. 8 shows that while patients did not improve at a statistically significant level, the median of average deviation was lesser with assistance across all shapes. This improvement is encouraging, and patients may require more refined assistance or multiple sessions to come to the level of a healthy subject in terms of average deviation.

Assistance did have a significant effect on patient speed ($p < 0.001$). Moreover, there was no difference between healthy subjects and patients with assistance when it came to their speeds ($p = 0.906$). This gives evidence that the platform's assistance is helping the patient move at speeds like healthy subjects. These results can be confirmed across all shapes in Fig. 8. The assistance helped improve the speed of patients when tracing the '∞'. The hexagon is more difficult to trace for patients as seen by the larger average deviation. While it is positive to see the assistance improving patients during their sessions, it is certainly not expected to completely reach and sustain the healthy standards within a single therapy session. Future work should include longitudinal studies to determine the effect of the platform over time.

# 5 Patient Classification and Trajectory Forecasting

While a subject is operating the RTM platform, it is important to realize their characteristics for providing accurate assistance. For this purpose, a DL model is trained which can:

i. classify if the person operating the RTM is healthy or unhealthy.
ii. forecast the movements of the subject's trajectory, so that assistance can be adjusted accordingly to mimic therapist's assistance.

A novel combination of one-dimensional convolution, Bi-directional Long Short-term Memory Network (BiLSTM) and self-attention mechanism are proposed for both the classification and forecasting tasks. These modules allow the model to capture local and long-term patterns through exploiting time series data. The classification model can provide a way to assist therapists in quantifying the patient's impairment, thereby tracking their improvement, and the forecasting model can facilitate independent therapy delivery. The data preprocessing, training, and performance comparison phase of the models are presented and detailed. The training of the model is performed using an 11th Gen Intel® Core™ i7-1160G7 CPU running at 1.20 GHz. The classification model takes around 1.38 minutes, while the forecasting model takes around 18.93 minutes.

## 5.1 Data Collection and Preprocessing

The data used in the experiments are time series $x \in \mathbb{R}^{n \times l}$ where $n = 3$ is the number of features, $l$ is the number of timesteps, and $\mathbb{R}$ is the set of real number. The features are the three stylus Cartesian coordinates. These were collected for each unassisted patient and subject for the various shapes. The triangle, 'B', and circle were used for training dataset, while the hexagon and '∞' are used for test dataset. The choice of splitting is based on the observation from previous results that hexagon being an advanced combination of circle and triangle and '∞' being a more complex 'B' with more curvature demanded. This could demonstrate the generalizability of the model from simple shapes to more complex ones. In both tasks, the same amount of data from both healthy subjects and patients is randomly selected to eliminate bias among training data. To unify the length of the training and test data, each series is truncated from 101st to 1000th sample. The offset in the starting point is intended to mitigate the instability upon starting.

## 5.2 Patient Classification

DL architectures can learn patterns and characteristics of the trace the RTM platform has collected. The trace is composed of thousands of temporal-ordered data points, which are perfect for sequential models such as Gated Recurrent Unit Network (GRU) [35], Long Short-term Memory Network (LSTM) [36], and the self-attention-based network [37]. Also, convolution can be utilized to capture local patterns. This intuition is further verified by applying isolation Forest (iForest) algorithm to detect outliers within trajectories of healthy subjects and patients [38]. The contamination rate is set to be 0.1, and it can be observed from Fig. 9 that outliers of patient trajectories are substantially more apparent. Therefore, the convolution layer needs to identify and emphasize local patterns (outliers), the BiLSTM layers need to capture temporal dependence, and the attention layer needs to subsequently assign more weights to the outliers, to carry out accurate predictions.

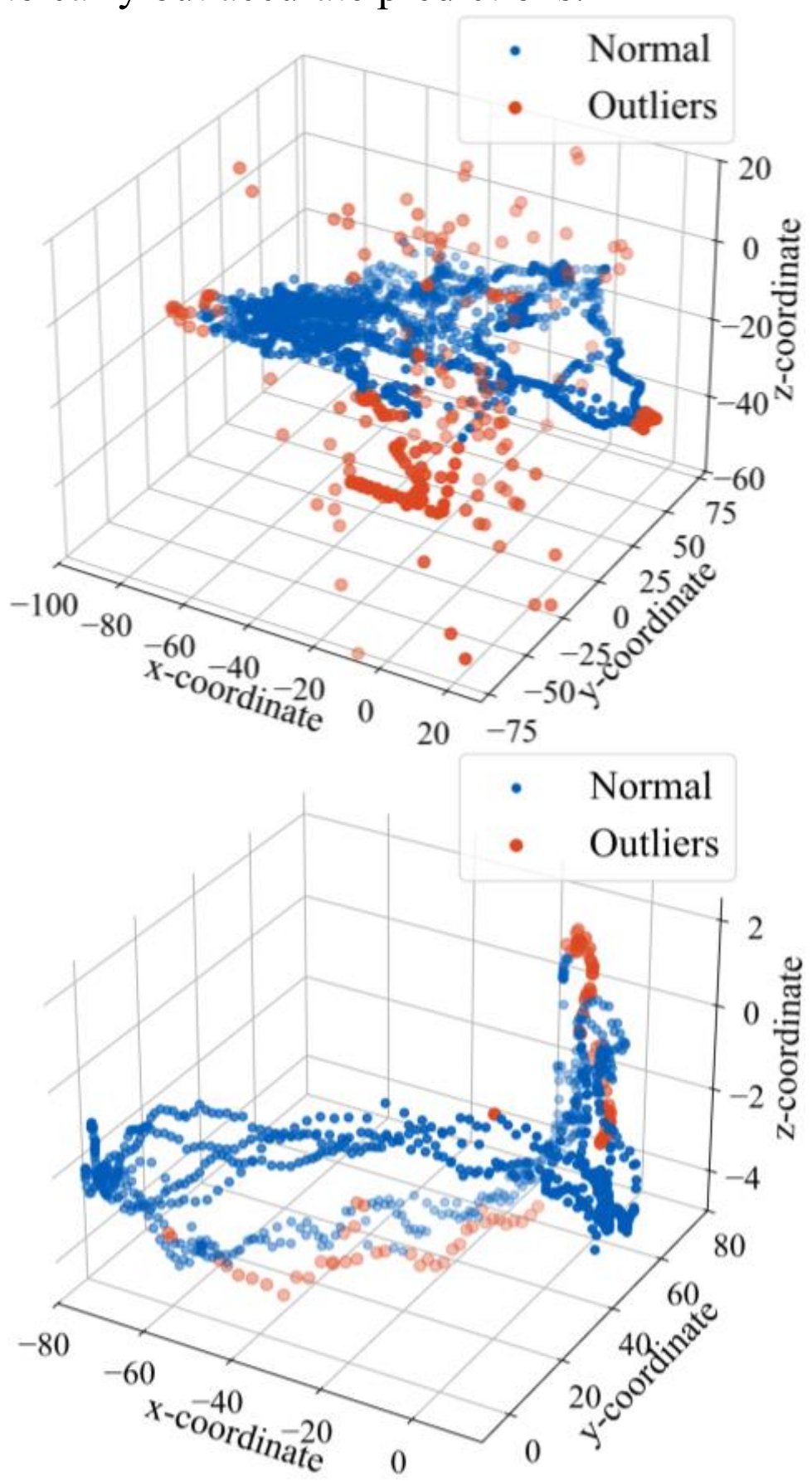

**Fig. 9** iForest results of one patient (top) and healthy subject (bottom) drawing "triangle"

Therefore, Conv@BiLSTM is proposed here. The architecture is first composed of one-dimensional convolution layer, followed by one-dimensional global maximum pooling (GMP) layer. The information is then passed to BiLSTM layers and one-head attention layer. Finally, the information goes through several fully connected layers to gradually de-scale. Since it is a binary classification task, it eventually scales down to a binary vector and is passed to a SoftMax activation function, which outputs the prediction result. Additionally, a Rectified Linear Unit (ReLU) activation function is applied after each layer, enhancing the performance of the model by introducing non-linearity. The architecture is shown in Fig. 10.

At the input layer, the data is passed with a size of $3 \times 900$ after trimming, which is then passed to the one-dimensional convolution layer, which adopts a kernel size of 3 with a default stride of 1, and the output channel size is set to be 64. Hence, the sliding convolution kernel with a size of 3 will roll over the entire time series to recognize local patterns, and a total of 64 key features will be output. The later-on GMP layer retains the most significant patterns while de-scaling the data for computation convenience since the stride here is set to 2, after which the data enters BiLSTM layers. BiLSTM consists of two LSTM networks with one going forward while the other going backward chronologically. The data is scaled up to 128 to contain more information.

These layers attempt to capture the temporal dependence of the subject's trajectory. Finally, the one-head attention layer aims to assign weights within the subject's trajectory, so that the model is informed which specific timesteps to concentrate. The residual fully connected layers just serve the role of gradually de-scaling to binary results, avoiding abrupt loss of information.

The performance of the proposed model is evaluated on 9 patients and 9 healthy subjects. 2 patients and 1 healthy subject were removed from training since those subjects did not complete all the shapes. Precision rate, F1-score, specificity rate, and accuracy rate are used as metrics. An ablation study is carried out to ensure that Conv@BiLSTM is the most effective architecture, by enabling various combinations of different layers (Table 3). GRU is also included in the comparison since it is essentially a simplified version of LSTM. Self-attention layers are denoted with "@". The proposed model performs best in all metrics. Self-attention and convolution improved the performance which is due to their capabilities to capture global and local relationships, respectively. It is interesting to note that BiLSTM performed better than GRU except when combined with self-attention, where the GRU architecture performed slightly better. The confusion matrix for the proposed architecture is presented in Fig. 11.

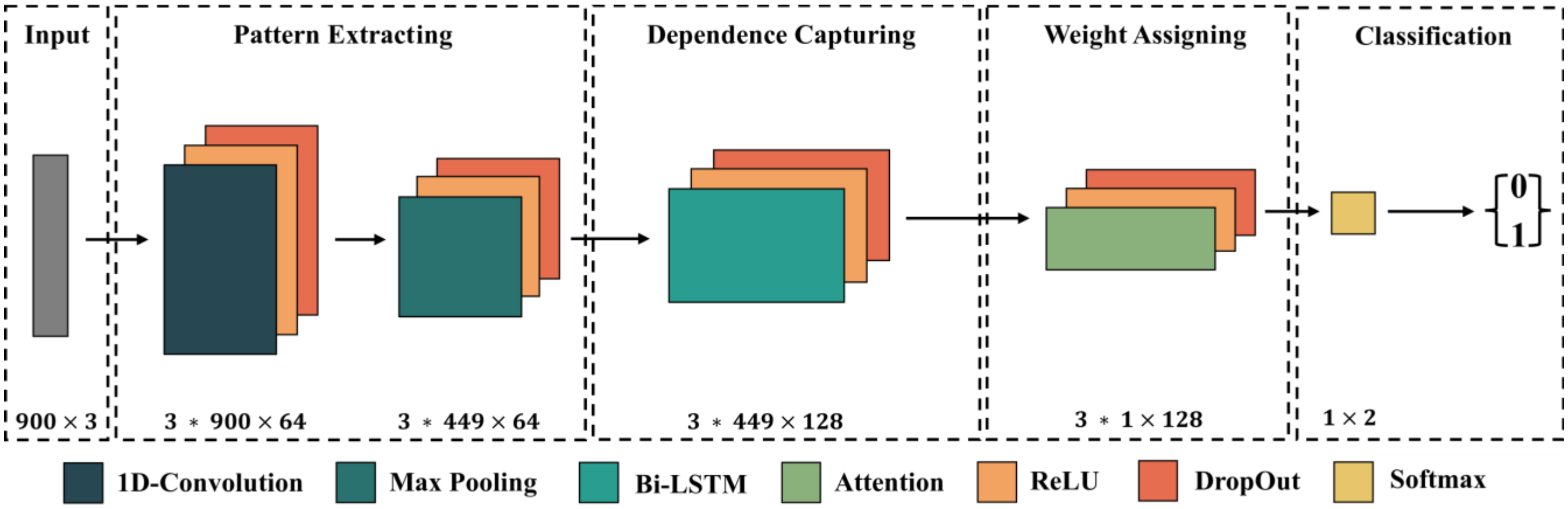

**Fig. 10** Simplified architecture of the proposed model

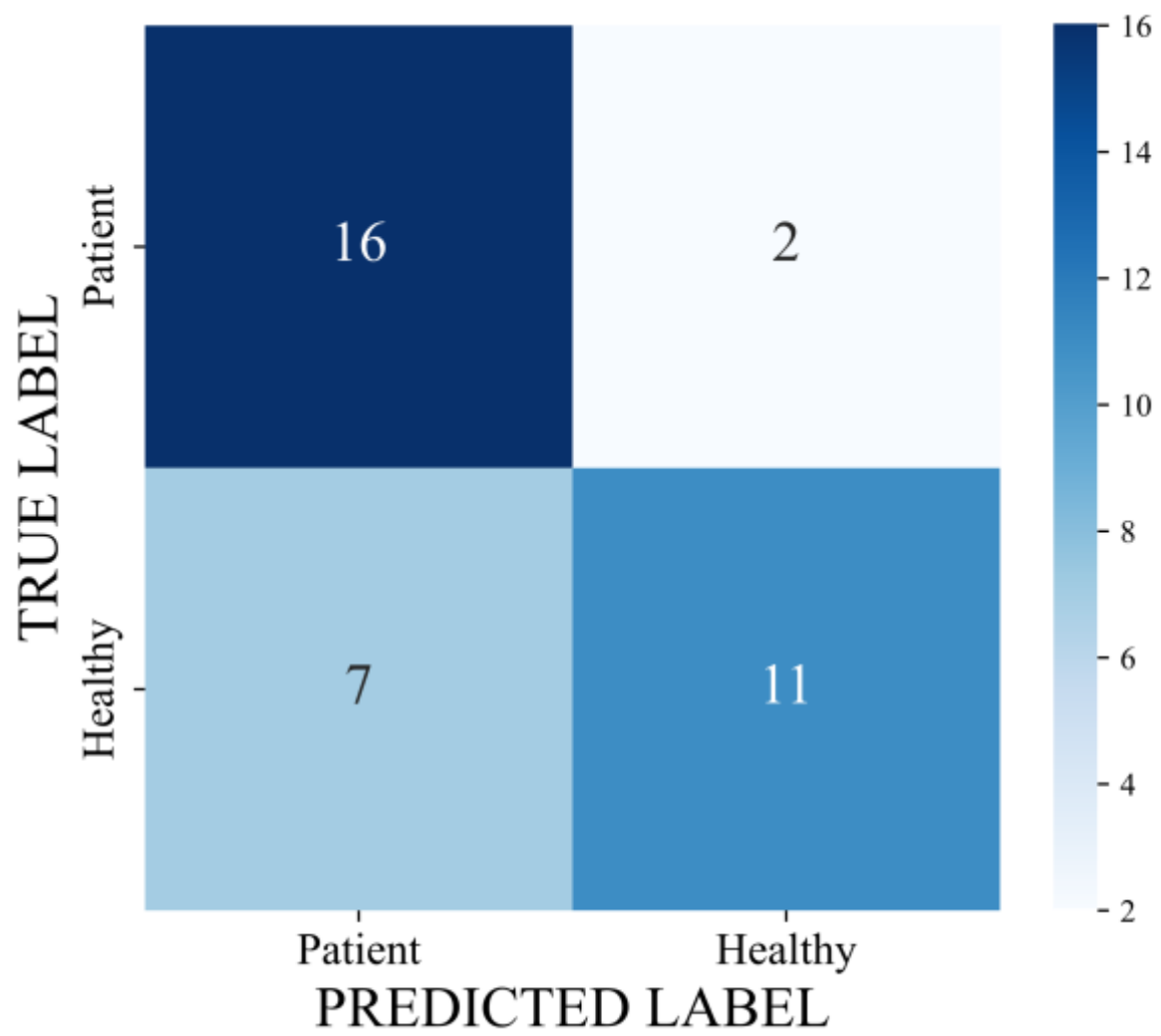

**Fig. 11** Confusion matrix of proposed model on classification task

**Table 3** Performance of classification models (**best in bold**)

| Model | Precision | F1 | Specificity | Accuracy |
|---|---|---|---|---|
| GRU | 0.556 | 0.476 | 0.636 | 0.528 |
| BiLSTM | 0.611 | 0.523 | 0.650 | 0.528 |
| @BiLSTM | 0.705 | 0.410 | 0.652 | 0.667 |
| @GRU | 0.722 | 0.703 | 0.545 | 0.667 |
| Conv@GRU | 0.813 | 0.703 | 0.667 | 0.528 |
| **Conv@BiLSTM (proposed)** | **0.889** | **0.711** | **0.846** | **0.750** |

## 5.3 Trajectory Forecasting

Forecasting the trajectory can be of great help for artificial intelligence-based rehabilitation, especially in terms of assisting the patient. Intuitively, therapists decide when to assist based on the patient's current trajectory and an empirical estimation of what the patient may do next. To account for the patient's past trajectory when predicting their future movements, fixed-length sequence-to-sequence models of patient trajectories are adopted. The data are split the same way as in Section 5.1 for both the training and test phase. Additionally, the previous 45 timesteps are used to forecast the next 5 timesteps, with a sliding window of size 5. Namely, if $\{a_n\}_{1\le n\le 900}$ is the test data after trimming, then $\{a_n\}_{5k-4\le n\le 5k+40}$ is the model input, and $\{a_n\}_{5k+41\le n\le 5k+45}$ is the ground truth output, where $k\epsilon[1,172]\wedge Z$ . Notably, the output is equivalent to $\{a_n\}_{46\le n\le 905}$ , which ensures temporal consistency in forecasting. Forecasting is run on the proposed model as in Fig. 10, with input size being $45 \times 3$. Also, a linear layer is applied after the attention layer to ensure the output size being $5 \times 3$.

Performance of forecasting is compared between the proposed model along with several state-of-the-art models including Encoder [37], Informer [39], GRU, and LSTM, and error metrics are introduced to compare the results, as shown in Table 4. Specifically, Mean Absolute Error (MAE), Mean Squared Error (MSE), Root Mean Squared Error (RMSE), and Mean Absolute Percentage Error (MAPE) are used as metrics. For short-time series forecasting, Conv@BiLSTM remains efficient for all metrics except for MAE. In Fig. 12 and Fig. 13, visualization is presented for forecasting using the proposed model. Such forecasting can predict the patients' movements when operating the RTM platform. Future work will explore how this forecasting capability can be incorporated to provide assistance. The proposed deep learning based predictive models and classifiers can further aid in providing customized experience to individual patients. For example, the models can be trained to separate tremors due to neurological conditions like Parkinson's disease from trajectory oscillations and inject impedance control and damping terms in addition to assistive forces to stabilize their motion.

**Table 4** Performance of forecasting models (**best in bold**)

| Model | MAE | MSE | RMSE | MAPE |
|---|---|---|---|---|
| Encoder | 5.06 | 52.16 | 7.22 | 77.07 |
| Informer | 5.33 | 68.43 | 8.27 | 121.08 |
| GRU | 5.29 | 35.73 | 5.32 | 89.13 |
| LSTM | **3.15** | 32.86 | 6.29 | 64.86 |
| Conv@GRU | 4.67 | 40.32 | 6.35 | 74.15 |
| **Conv@BiLSTM (proposed)** | 3.37 | **23.31** | **4.83** | **63.69** |

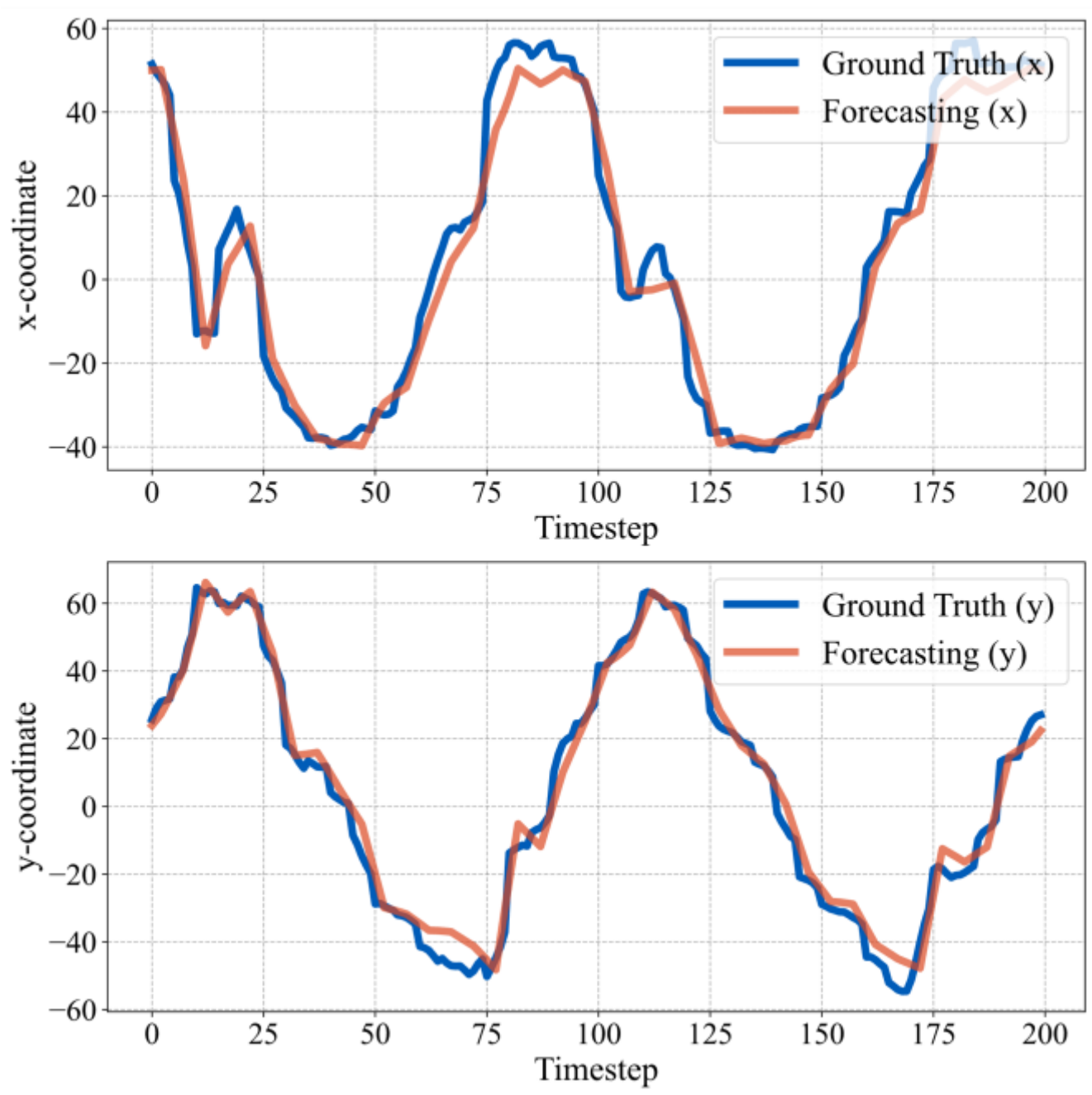

**Fig. 12** Temporal comparison of hexagon foresting result and ground truth for proposed model

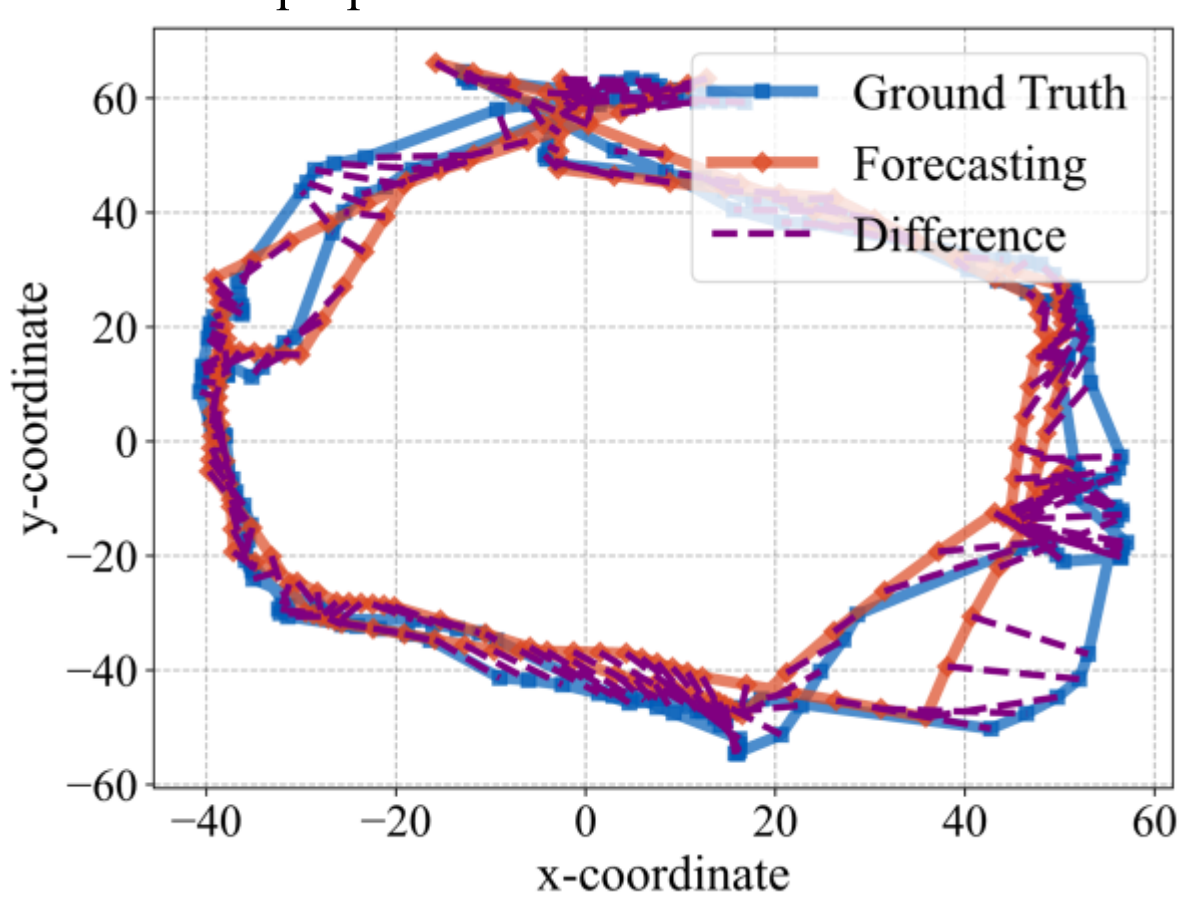

**Fig. 13** Spatial comparison of hexagon prediction result and ground truth for proposed model with their differences

# 6 Conclusions

A robotic version of the traditional trail making test has been developed in this work. The platform targets to simultaneously assess and assist the patients with variegated neurological conditions, as well as to minimize the workload of the therapists. With all assistance, safety and data recording features incorporated, the developed robotic trail maker is validated through testing with patients and healthy subjects. The platform was setup at OSF Outpatient Rehabilitation Center, Peoria, IL, where 10 healthy and 11 patients participated in the study. Results demonstrate that the robotic assistance improves the patients' metrics such as accuracy and speed, bringing them closer to healthy subjects' performance. Statistical analysis and inferences from the metrics were discussed. A deep learning architecture is proposed for utilizing the collected data for forecasting and classification. Future work would deploy this model for tracking patient progress and autonomous delivery of personalized assistance. In summary, the proposed platform is demonstrated to intelligently assist the patients by evaluating their performance while using the platform.

**Funding** This work has been funded by the Jump ARCHES program, a collaborative effort between the Health Care Engineering Systems Center, University of Illinois Urbana Champaign, and the OSF Health Care, Peoria, IL, under the grant P-386. Nadica Miljković was partly supported by the Ministry of Science, Technological Development and Innovation of the Republic of Serbia under contract No. 451-03-137/2025-03/200103.

# Statements and Declarations

**Ethics Approval and Consent to Participate** The methodology for this study was approved by the Peoria Institutional Review Bord (IRB #00000688).

**Competing Interests** The authors declare to have no competing interests.

**Data Availability** The code for implementation of the RTM platform can be found at [40], and the data and code for data analysis is available at [41].